\documentclass{article}
\usepackage{spconf,amsmath,graphicx,hyperref}
\usepackage{tabu}                      
\usepackage{booktabs}                  
\usepackage{lipsum}                    
\usepackage{mwe}                       
\usepackage{multirow}
\usepackage{mathptmx}                  
\usepackage{bm}

\usepackage{xspace}
\makeatletter
\DeclareRobustCommand\onedot{\futurelet\@let@token\@onedot}
\def\@onedot{\ifx\@let@token.\else.\null\fi\xspace}
\def\eg{\emph{e.g}\onedot} 
\def\ie{\emph{i.e}\onedot}

\def\etal{\emph{et al}\onedot}
\makeatother

\newcommand{\Tref}[1]{Table~\ref{#1}}

\newcommand{\fref}[1]{Fig.~\ref{#1}}

\usepackage{xcolor}
\newcounter{todos}
\AtEndDocument{\ifnum\value{todos}>0 \PackageWarning{TODOS}{There are \arabic{todos} todos left in this paper! Fix them before submitting the paper!} \fi}

\newcommand{\real}{\mathbb{R}}

\usepackage{bm}

\newcommand{\diag}{\text{diag}}

\usepackage{mathtools}
\usepackage{amsmath,amssymb}

\title{Gaussian Mesh Renderer for Lightweight Differentiable Rendering}
\name{Xinpeng Liu, Fumio Okura\thanks{This work was partly supported by JSPS KAKENHI JP23H05491, JP25K03140, JST ASPIRE JPMJAP2502, and JST FOREST JPMJFR206F.}}
\address{The University of Osaka}

\begin{document}
%
\maketitle
\begin{abstract}
    3D Gaussian Splatting (3DGS) has enabled high-fidelity virtualization with fast rendering and optimization for novel view synthesis. On the other hand, triangle mesh models still remain a popular choice for surface reconstruction but suffer from slow or heavy optimization in traditional mesh-based differentiable renderers. To address this problem, we propose a new lightweight differentiable mesh renderer leveraging the efficient rasterization process of 3DGS, named Gaussian Mesh Renderer (GMR), which tightly integrates the Gaussian and mesh representations. Each Gaussian primitive is analytically derived from the corresponding mesh triangle, preserving structural fidelity and enabling the gradient flow. Compared to the traditional mesh renderers, our method achieves smoother gradients, which especially contributes to better optimization using smaller batch sizes with limited memory. Our implementation is available in the public GitHub repository (\url{https://github.com/huntorochi/Gaussian-Mesh-Renderer}).\looseness=-1
\end{abstract}

\section{Introduction}
\label{sec:intro}
\vspace{-2mm}
3D reconstruction of real-world objects through gradient-based optimization (\ie, inverse rendering) is becoming fundamental to real-world virtualization~\cite{orignerf,Neus,volSDF}. Especially, 3D Gaussian Splatting (3DGS)~\cite{orig3dgs} has recently emerged as an efficient way for real-time novel view synthesis. By representing a scene as a collection of anisotropic Gaussians, 3DGS supports parallel rasterization and provides analytical gradients, enabling real-time rendering even on consumer hardware. While recent works like SuGaR~\cite{sugar} and 2DGS~\cite{2dgs} introduce constraints into 3DGS to improve surface consistency, 3DGS-based methods still lack explicit structural constraints, resulting in limited applicability to geometry-sensitive tasks such as physics simulation or scene editing.

Mesh models, on the other hand, provide compact, explicit surface representations naturally compatible with standard graphics pipelines. They offer strong geometric priors and essentially support high-quality rendering with high-polygon mesh models. Traditional mesh rendering processes determine which pixels lie inside projected geometric primitives (\ie, rasterization). Since this process uses binary edge tests of the mesh boundary, commonly implemented using a hard step function~\cite {hardRas}, the rasterization process naturally introduces non-differentiability at mesh boundaries.\looseness=-1 

To incorporate the mesh representations for gradient-based optimization, differentiable mesh rendering~\cite{openDR,NMR,kato2020differentiable} has been actively studied. 
A popular formulation proposed in Soft Rasterizer (SoftRas)~\cite{softRas} replaces hard step functions of mesh boundary with smooth probability functions to make mesh rendering differentiable. SoftRas has become the \emph{de facto} standard in differentiable mesh rendering, which is a basis of many variants~\cite{expRas,gausRas,Pix2VexIR}. Laine~\etal~\cite{Nvdiffrast} further propose a GPU-efficient implementation, Nvdiffrast, computing coverage based on edge-crossing between neighboring pixels. However, these local formulations inherently restrict gradient flow to triangle boundaries, thus limiting their expressiveness for global shape or visibility reasoning. This drawback severely affects optimization with a smaller batch size, making it more likely to become unstable.\looseness=-1
   
Some recent studies explore combining mesh priors into 3DGS. In \cite{games, gaussianavatars, MeshGaussian}, Gaussians are initialized at mesh triangle centroids, assigning orientation and scale through different strategies. 
Similarly, \cite{Direct3DMesh, Dynamic3DMesh, MeshGS} optimizes Gaussians and mesh surfaces separately, then aligns them through a nearest-neighbor manner or correspondence constraints. 
While both approaches offer partial structural guidance to 3DGS, they often lack structural connectivity among polygons or require additional supervision, making them difficult to use for differentiable mesh rendering purposes. 

On top of the recent efforts to combine mesh priors with 3DGS, we propose a lightweight differentiable rendering method for mesh models leveraging 3DGS's rasterizer. 
By analytically converting meshes to Gaussian primitives, we yield a simple differentiable mesh rasterizer, hereafter called the Gaussian Mesh Renderer (GMR), which assumes a fixed mesh topology and supports end-to-end optimization via gradient flow from rendered images to underlying mesh geometry and thus can be directly used as an alternative to traditional differentiable mesh renderers such as SoftRas.\looseness=-1

Experiments show that our GMR allows for memory-efficient and accurate surface reconstruction, especially in the optimization using small batch sizes, compared with existing mesh rasterizers and 3DGS-based methods. The results imply the suitability of our approach for optimizing mesh models in lightweight applications, such as in mobile environments.

\section{Method: Gaussian Mesh Renderer (GMR)}
\vspace{-2mm}
Given a mesh, GMR analytically converts each facet into one anisotropic Gaussian whose support lies on the facet plane. Unlike ordinary 3DGS, Gaussians are constrained to be (nearly) planar and \emph{geometrically aligned} with the source facet, which provides strong structural priors. \looseness=-1

\vspace{-2mm}
\subsection{3D Gaussian Splatting (3DGS)}
\vspace{-2mm}
We follow the standard 3DGS representation and differentiable rasterizer~\cite{orig3dgs}. Each primitive is a Gaussian with 3D mean $\bm{\mu}\in\real^3$ and a covariance matrix $\bm{\Sigma}\in\real^{3\times3}$:
\begin{equation}
  {G}(\bm{p}) = \exp \left(-\frac{1}{2} (\bm{p}-\bm{\mu})^\top \bm{\Sigma}^{-1} (\bm{p}-\bm{\mu})\right),
  \vspace{-1.8mm}
\end{equation}
where $\bm{p}\in\real^3$ denotes a point in the 3D Cartesian coordinates. 
The covariance matrix $\bm{\Sigma}$ is decomposed into a scaling matrix $\bm{S}=\diag(s_1,s_2,s_3) \in\real_{\ge0}^3$ and a rotation matrix $\bm{R}\in \mathrm{SO}(3)$ as
\begin{equation}
  \bm{\Sigma} = \bm{R}\bm{S}\bm{S}^\top\bm{R}^\top.
  \vspace{-1.8mm}
\end{equation}

In addition to the scale and rotation, 3DGS pipeline optimizes Gaussian appearances $\{o,\bm{c}\}$ containing opacity $o$ and color $\bm{c}$ represented as its RGB direct component (DC) and view-dependent spherical harmonics coefficients.
During the rendering, 3D Gaussians are projected onto the 2D camera image planes in a differentiable \emph{splatting} process~\cite{3dgsFather}.

\vspace{-2mm}
\subsection{Mesh-to-Gaussian Conversion}
\vspace{-2mm}
\label{sec:mesh2gaussian}
Let a triangle mesh as $\mathcal{M}=(\mathcal{V}, \mathcal{F})$, where vertices $\mathcal{V}=\{\bm{v}_1, ..., \bm{v}_n\}\in\real^3$ and facets $\mathcal{F}=\{f_1, ..., f_m\}$ with each $f$ denoting a triplet of vertex indices. Our goal is to convert each facet into a Gaussian $\bigl(\bm{\mu}_f, \bm{\Sigma}_f\bigr)$ that faithfully approximates the original surface geometry. 

\vspace{1.6mm}
\noindent
\textbf{Local Coordinate System on a Facet.}
For each triangular facet defined by vertices $(\bm{v}_i,\bm{v}_j,\bm{v}_k)\in\real^3$, we define a local orthonormal coordinate system embedded on the facet plane. Specifically, we place the origin at $\bm{v}_i$, and let the $x$-axis point in the direction of $\bm{v}_j - \bm{v}_i$. The $y$-axis is chosen to lie in the facet plane and be orthogonal to the $x$-axis; in practice, it can be computed via Gram-Schmidt orthogonalization using $\bm{v}_k - \bm{v}_i$. The facet normal $\bm{n}_f$ serves as the implicit $z$-axis of this local frame and is directly available from the mesh structure. 

Any point $\bm{p}\in\real^3$ on this facet can then be represented in the local frame as
\begin{equation}
   \bm{p} 
   \;=\;
   \bm{v}_i
   \;+\;
   x\,\bm{x}_{\text{axis}}
   \;+\;
   y\,\bm{y}_{\text{axis}},
   \label{eq:2d_local_coords}
   \vspace{-1.8mm}
\end{equation}
where $(x,y)\in\real^2$ are the local coordinates with respect to $\bm{v}_i$.

\vspace{1.6mm}
\noindent
\textbf{2D Covariance in the Local Plane.}
We compute a 2D Gaussian that approximates the uniform distribution on the triangle formed by $(\bm{v}_i,\bm{v}_j,\bm{v}_k)$ in the local $xy$ coordinate system defined above. Let the projected 2D positions of the three vertices be
\[
  \bm{v}_i^{\text{(2D)}} = [0,0]^\top,\quad
  \bm{v}_j^{\text{(2D)}} = [x_j,0]^\top,\quad
  \bm{v}_k^{\text{(2D)}} = [x_k, y_k]^\top,
\]
where $\bm{v}_j$ lies on the $x$-axis and $\bm{v}_k$ is on the $(x,y)$ plane. 

The centroid of the triangle is located at
\begin{equation}
  [\,\mu_x,\,\mu_y\,]^\top
  \;=\;
  \tfrac{1}{3}\left[
    x_j + x_k,\;
    y_k
  \right]^\top,
  \vspace{-1.8mm}
\end{equation}
and we can represent any point inside the triangle using barycentric coordinates $(\alpha, \beta) \in [0,1]$ with $\alpha + \beta \leq 1$:
\begin{align}
  x(\alpha,\beta) 
  &= x_i + \alpha(x_j - x_i) + \beta(x_k - x_i), \nonumber\\
  y(\alpha,\beta) 
  &= y_i + \alpha(y_j - y_i) + \beta(y_k - y_i).
  \label{eq:barycentric_coords}
  \vspace{-1.8mm}
\end{align}

The second-order moments of the uniform distribution over the triangle are given by:
\begin{align}
  \mathbb{E}[x^2]
  &= \frac{1}{|A|} \iint_{\triangle} x(\alpha,\beta)^2\, dA,
  \label{eq:ex2_integral} \\
  \mathbb{E}[xy]
  &= \frac{1}{|A|} \iint_{\triangle} x(\alpha,\beta)y(\alpha,\beta)\, dA,
  \label{eq:exy_integral} \\
  \mathbb{E}[y^2]
  &= \frac{1}{|A|} \iint_{\triangle} y(\alpha,\beta)^2\, dA,
  \label{eq:ey2_integral}
  \vspace{-1.8mm}
\end{align}
where $A$ and $|A|$ respectively denote the triangle's area and size on its plane.
Since the integrands are low-degree polynomials in $(\alpha, \beta)$, all integrals above admit closed-form solutions, avoiding any sampling. 
The 2D covariance matrix in the local coordinate system is then constructed as:
\begin{equation}
  \bm{\Sigma}^{(2D)} 
  \;=\;
  \begin{bmatrix}
    \mathbb{E}[x^2] - \mu_x^2 
    & 
    \mathbb{E}[xy] - \mu_x\,\mu_y
    \\
    \mathbb{E}[xy] - \mu_x\,\mu_y
    & 
    \mathbb{E}[y^2] - \mu_y^2
  \end{bmatrix}.
  \label{eq:cov_2d}
  \vspace{-1.8mm}
\end{equation}
When $|A|>0$, \ie, three vertices are not collinear, the resulting $\bm{\Sigma}^{\text{(2D)}}$ is guaranteed to be positive semi-definite. 

\vspace{1.6mm}
\noindent
\textbf{Lifting 2D Covariance to 3D.}
To convert the 2D covariance into a 3D anisotropic Gaussian, we embed it into the original 3D space defined by the local orthonormal frame $(\bm{x}_{\text{axis}}, \bm{y}_{\text{axis}}, \bm{n}_f)$. 
Let $\bm{R}_f \in \real^{3\times3}$ be the rotation matrix whose columns are these three unit vectors, and denote the eigen decomposition of the 2D covariance matrix $\bm{\Sigma}^{\text{(2D)}}$ as
\begin{equation}
  \bm{\Sigma}^{\text{(2D)}} = \bm{U} \cdot \mathrm{diag}(\lambda_1, \lambda_2) \cdot \bm{U}^\top,
  \vspace{-1.8mm}
\end{equation}
where $\lambda_1 \ge \lambda_2$ are the eigenvalues and $\bm{U} = [\bm{u}_1, \bm{u}_2]$ contains the corresponding principal directions in the 2D plane.

We first lift the 2D eigenvectors into 3D as
\begin{align}
  \bm{x}'_{\text{3D}} & u_{1x} \bm{x}_{\text{axis}} + u_{1y} \bm{y}_{\text{axis}}, \quad \bm{y}'_{\text{3D}} = u_{2x} \bm{x}_{\text{axis}} + u_{2y} \bm{y}_{\text{axis}},
  \vspace{-1.8mm}
\end{align}
where $\bm{u}_1 = [u_{1x}, u_{1y}]^\top$ and $\bm{u}_2 = [u_{2x}, u_{2y}]^\top$.

To ensure geometric consistency, we match the area of the Gaussian's $1$-sigma ellipse, $\pi \sqrt{\lambda_1 \lambda_2}$, to the original triangle. 
Given triangle size ${|A|}$, the scaling factor $\kappa$ and the rescaled eigenvalues (\ie, Gaussian's scale parameters) are
\begin{equation}
  \kappa = \frac{|A|}{\pi \sqrt{\max(\lambda_1 \lambda_2,\, \varepsilon)}}, \quad s_x^2 = \kappa \lambda_1, \quad s_y^2 = \kappa \lambda_2,
  \vspace{-1.8mm}
\end{equation}
where we set $\varepsilon=10^{-14}$ for numerical stability under near-degenerate triangles. We set $s_z = 10^{-6}$ to represent the flat surfaces of the meshes.

Finally, we construct the 3D covariance matrix of the Gaussian using these rescaled in-plane scales:
\begin{equation}
  \bm{\Sigma}_f^{\text{(3D)}}
  \;=\;
  \bm{R}_{\text{local}}
  \cdot
  \mathrm{diag}\left(
    s_x^2,\, s_y^2,\, s_z^2
  \right)
  \cdot
  \bm{R}_{\text{local}}^\top,
  \label{eq:cov_3d_diag}
  \vspace{-1.8mm}
\end{equation}
where the rotation matrix $\bm{R}_{\text{local}} = [\bm{x}'_{\text{3D}}, \bm{y}'_{\text{3D}}, \bm{n}_f]$ aligns the Gaussian with the principal directions of $\bm{\Sigma}^{\text{(2D)}}$.

Note that $\bm{R}_{\text{local}}$ is generally \emph{not} equal to $\bm{R}_f$, as the in-plane eigenvectors $\bm{U}$ may introduce a rotation within the local frame. This transformation can be expressed as
\begin{equation}
  \bm{R}_{\text{local}} 
  \;=\;
  \bm{R}_f
  \begin{bmatrix}
    \bm{U} & \bm{0} \\
    \bm{0}^\top & 1
  \end{bmatrix},
  \label{eq:Rlocal_vs_Rf}
  \vspace{-1.8mm}
\end{equation}
which rotates the canonical local frame $\bm{R}_f$ by the in-plane principal orientation. This ensures that the resulting 3D Gaussian is both geometrically aligned with the facet and statistically aligned with its mass distribution.

Given that $s_z^2 \approx 0$, an alternative formulation to Eq.~\eqref{eq:cov_3d_diag} can be derived by directly embedding the original 2D covariance matrix into 3D as
\begin{equation}
  \bm{\Sigma}_f^{\text{(3D)}} 
  \;=\;
  \bm{R}_f
  \begin{bmatrix}
    \bm{\Sigma}^{\text{(2D)}} & \bm{0} \\
    \bm{0} & 0
  \end{bmatrix}
  \bm{R}_{f}^\top.
  \label{eq:cov_3d_embed}
  \vspace{-1mm}
\end{equation}
Our implementation uses this form to emphasize that the distribution is strictly planar and aligns precisely with the principal axes of the 2D geometry.

\begin{table}[t]
    \centering
 \resizebox{\linewidth}{!}{
    \begin{tabular}{c|cc|ccc} 
    \toprule
\multirow{2}{*}{Method / Metric} & \multicolumn{2}{c|}{Geometric accuracy} & \multicolumn{3}{c}{Rendering quality}\\ 
            & CD $\downarrow$ & NC $\uparrow$ & PSNR $\uparrow$ & SSIM $\uparrow$ & LPIPS $\downarrow$  \\ \midrule
SoftRas~\cite{softRas}
& $3.10 \times 10^{-3}$/\;\;\;\;\;\;\;---\;\;\;\;\;\;\;   & 0.615/\;\;---\;\;  & 22.32/\;\;---\;\;  & 0.890/\;\;---\;\;  & 0.117/\;\;---\;\; \\
Nvdiffrast~\cite{Nvdiffrast}
& $8.02 \times 10^{-5}$/$1.99 \times 10^{-5}$         & 0.927/0.933        & 26.76/27.32   & 0.949/0.959   & 0.051/\textbf{0.043}\\
3DGS-based
& $1.12 \times 10^{-2}$/$9.41 \times 10^{-3}$         & 0.046/0.332          & 20.56/23.24        & 0.794/0.859   & 0.159/0.134 \\
GMR (Ours)
& $\bm{1.54 \times 10^{-5}}$/$\bm{1.67 \times 10^{-5}}$ & \textbf{0.965}/\textbf{0.967} & \textbf{27.43}/\textbf{27.59}   & \textbf{0.959}/\textbf{0.962}   & \textbf{0.047}/0.044\\
\bottomrule
    \end{tabular}
    }
    \vspace{-3mm}
    \caption{Quantitative results in a \{batch size 1\}/\{batch size 10\} format. The best scores are marked \textbf{bold}. }
    \label{tab:quantitative}
    \vspace{-2mm}
\end{table}

\vspace{1.6mm}
\noindent
\textbf{Color Assignment for Gaussians.}
To complete the definition of each 3D Gaussian, we assign opacity $o$ and color $\bm{c}$ to every Gaussian based on its originating mesh facet. 
In our work, we set the opacity $o = 1$ for all Gaussians, modeling them as fully opaque, solid components.
Since the underlying mesh typically includes per-vertex color attributes or texture mappings, we define the Gaussian RGB color $\bm{c}_f \in \real^3$ as the average color of its three vertices:
\begin{equation}
  \bm{c}_f = \tfrac{1}{3} \left(\bm{c}_i + \bm{c}_j + \bm{c}_k\right),
\end{equation}
where $\bm{c}_i, \bm{c}_j, \bm{c}_k \in \real^3$ are the RGB values at vertices $(\bm{v}_i,\bm{v}_j,\bm{v}_k)$. 
The color representation can be straightforwardly extended to non-Lambertian (\ie, view-dependent) surfaces using mesh's physically-based rendering (PBR) materials and Gaussian's full spherical harmonics representations, while we now implement the conversion of base RGB color assuming Lambertian surfaces for simplicity, and thus $\bm{c}_f$ is stored onto Gaussian's DC component in the color attribute $\bm{c}$.

\begin{figure*}[t]
    \centering
    \includegraphics[width=\linewidth]{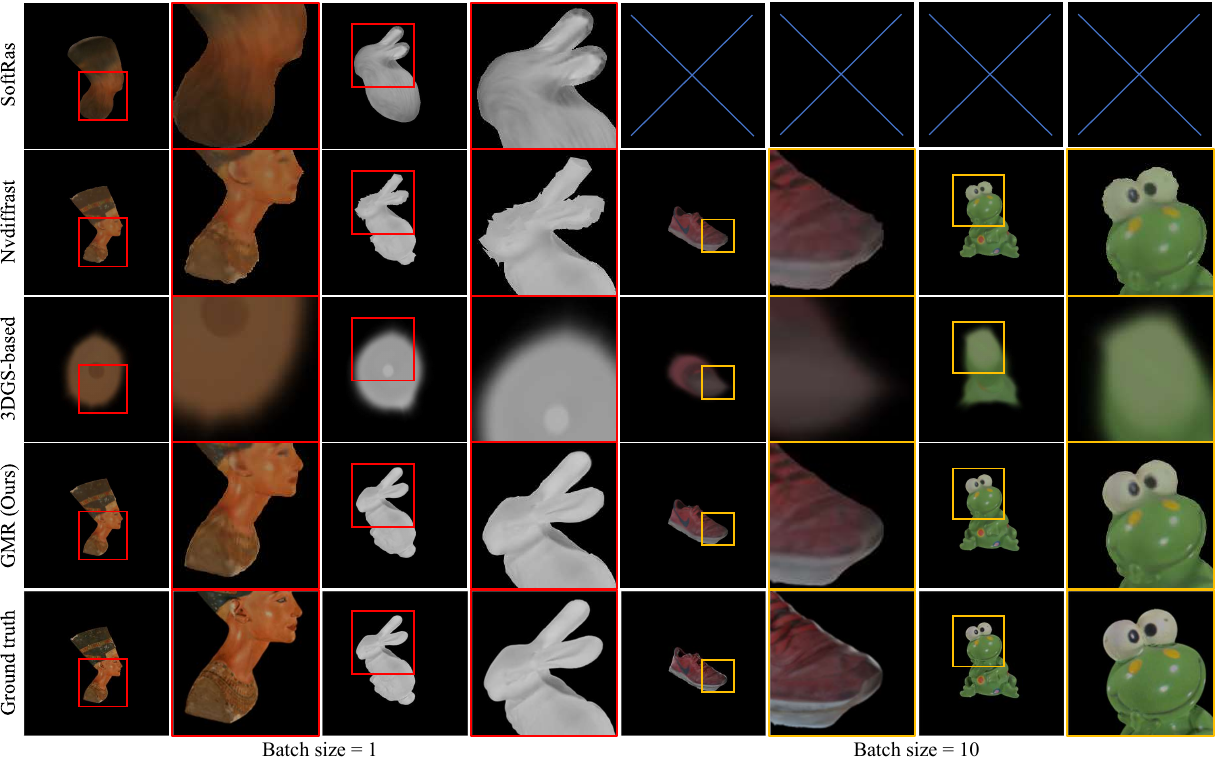}
    \caption{Visual results for batch size $=1$ and batch size $=10$. The PyTorch3D's SoftRas implementation does not support mini-batch inputs during the optimization.} 
    \label{fig:visual}
\end{figure*}

\section{Experiments}

\vspace{-2mm}
\subsection{Experiment Setup}
\vspace{-2mm}

We compare two typical cases of optimization: Using a single image for an iteration (\ie, batch size $=1$) and mini-batch-based optimization (\ie, batch size $=10$). 
These small batch sizes mimic the optimization in small GPUs with limited memory (\eg, in mobile devices), and are commonly used practically; for example, a well-known SoftRas~\cite{softRas} implementation in PyTorch3D only supports the batch size of $1$.

\vspace{1.6mm}
\noindent\textbf{Implementation Details.}
All experiments were conducted on an NVIDIA RTX 4090 GPU. Our renderer is implemented in PyTorch and uses the gsplat rasterizer~\cite{gsplat}. Unless otherwise stated, optimization starts from the same normalized sphere (unit radius, centered at the origin), tessellated into approximately 40K triangular facets. We adopt VectorAdam~\cite{vectoradam} as the default optimizer for all experiments. 

\vspace{1.6mm}
\noindent
\textbf{Loss Functions.} All methods are optimized with the same set of losses and their weights across all scenes and methods, which are common reconstruction and mesh regularization losses for mesh optimization. 
\emph{Color loss} minimizes the mean squared error (MSE) between the rendered and the ground-truth images.
\emph{Silhouette loss} measures binary cross-entropy between the rendered mask and the ground-truth object silhouette.
\emph{Edge length loss} encourages uniform edge lengths and suppresses artifacts.
\emph{Laplacian smoothing loss} promotes mesh smoothness by minimizing local geometric distortion.\looseness=-1

\vspace{1.6mm}
\noindent
\textbf{Dataset.}
We use $17$ objects selected from the Common 3D Test Models~\cite{common} and Objaverse~\cite{Objaverse} datasets, including both untextured and textured objects. In all our experiments, we adopt the same rendering process to generate 253 views for each 3D model, obtained from uniformly sampled camera viewpoints covering a hemisphere around the object.

\vspace{1.6mm}
\noindent
\textbf{Metrics.}
We compare geometric accuracy after optimization using \emph{chamfer distance (CD)} between the predicted and the ground-truth surfaces, and \emph{normal consistency (NC)} measuring the angle deviation between corresponding surface normals. 
We also assess novel-view rendering performance using common image-space metrics, \ie, \emph{PSNR}, \emph{SSIM}, and \emph{LPIPS}. 

\vspace{1.6mm}
\noindent
\textbf{Baselines.}
We compare our method with state-of-the-art differentiable rendering baselines. For differentiable mesh rasterizers, we use PyTorch3D's Soft Rasterizer (\textbf{SoftRas})~\cite{softRas} and Nvidia's \textbf{Nvdiffrast}~\cite{Nvdiffrast}. We also straightforwardly implement the state-of-the-art mesh-aware \textbf{3DGS-based} methods for the mesh rendering purpose, following the GaussianAvatars~\cite{gaussianavatars} and GaMeS~\cite{games}. Specifically, we locate each Gaussian at the triangle center and compute rotation using the triangle normal and the first edge direction, where we set the $z$-axis to the triangle normal and the $x$-axis to the first edge direction. This approach resembles a simplified Gram-Schmidt process but lacks geometric guarantees for accurate alignment with the triangle's shape. 

\subsection{Results}
\noindent
\textbf{Main Results.}
\Tref{tab:quantitative} shows the average quantitative scores for all 3D models; and \fref{fig:visual} shows visual examples. 
In both setups of the batch sizes of $1$ and $10$, our GMR achieves the best accuracy thanks to its smoother gradient.
Using existing mesh rasterizers often produces visible artifacts, such as jagged boundaries or incomplete shape recovery, from the same initialization.
The straightforward implementation of existing 3DGS-based methods yields significantly worse accuracy than ours. The coordinate frames derived from edge and normal directions fail to capture the actual geometry of the triangle, making it difficult for the optimization to update the mesh vertex positions effectively.

\vspace{1.6mm}
\noindent
\textbf{Performance.} At batch size $=10$, compared to Nvdiffrast, our method uses around 1,000~[MB] peak GPU memory and reduces peak memory by $\sim$30\% under the same settings, and converges in 5,000--15,000 iterations, which is affordable for recent mobile devices.
At batch size $=1$, our method is $\sim$40\% faster than SoftRas under an equivalent PyTorch setup.

\vspace{1.6mm}
\noindent
\textbf{Robustness to Initialization.} To further analyze the behaviors under tricky initialization conditions, we shift the initial mesh center away from the object's center by $0.5$, $1.0$, and $1.5$, where all objects are normalized such that the maximum distance between any two vertices is $2.0$. Our method consistently converges to consistent geometry, even from far-off initializations, demonstrating strong robustness to spatial misalignment of initial shape.

\section{Conclusion}
This paper introduces a lightweight differentiable mesh renderer, GMR, which tightly combines the mesh representation and the 3DGS's fast rasterizer. 
Unlike existing differentiable renderers, our method robustly works regardless of initialization, even with small batches and low memory consumption, thanks to smoother gradients.

\vspace{1.6mm}
\noindent
\textbf{Limitation.}
While GMR achieves better accuracy, it still runs slower than Nvdiffrast's well-optimized GPU implementation~\cite{Nvdiffrast}. 
Especially, using large batch sizes ($> 50$), Nvdiffrast shows both better accuracy and faster optimization compared to ours. We have not developed a CUDA implementation for the mesh-to-Gaussian conversion process; viable future work is to implement the process on a GPU and combine it with the 3DGS's rasterizer to improve throughput.

\clearpage

\bibliographystyle{IEEEbib}
\bibliography{main_paper}

\end{document}